\documentclass{article}
\usepackage{spconf,amsmath,epsfig,xcolor,subcaption,graphicx,multirow,csquotes}
\usepackage{url}

\title{Shining a Light on Hurricane Damage Estimation via Nighttime Light Data: Pre-processing Matters}

\name{\begin{tabular}{c}
Nancy Thomas \sthanks{Starred authors had equal contributions. \\ \copyright 2024 IEEE.  Personal use of this material is permitted. 
Permission from IEEE must be obtained for all other uses, in any current or future media, 
including reprinting/republishing this material for advertising or promotional purposes, 
creating new collective works, for resale or redistribution to servers or lists, 
or reuse of any copyrighted component of this work in other works.} \qquad Saba Rahimi \textsuperscript{*} \\
Annita Vapsi \qquad Cathy Ansell \qquad Elizabeth Christie \\
Daniel Borrajo \qquad Tucker Balch \qquad Manuela Veloso
\end{tabular}}

\address{J.P. Morgan Chase \& Co. AI Research and Climate Risk}

\address{J.P. Morgan Chase \& Co. AI Research and Climate Risk}

\begin{document}
%
\maketitle
\begin{abstract}
Amidst escalating climate change, hurricanes are inflicting severe socioeconomic impacts, marked by heightened economic losses and increased displacement. Previous research utilized nighttime light data to predict the impact of hurricanes on economic losses. However, prior work did not provide a thorough analysis of the impact of combining different techniques for pre-processing nighttime light (NTL) data. Addressing this gap, our research explores a variety of NTL pre-processing techniques, including {\it value thresholding}, {\it built masking}, and {\it quality filtering and imputation}, applied to two distinct datasets, VSC-NTL and VNP46A2, at the zip code level. Experiments evaluate the correlation of the denoised NTL data with economic damages of Category 4-5 hurricanes in Florida. They reveal that the quality masking and imputation technique applied to VNP46A2 show a substantial correlation 
with economic damage data.

\end{abstract}

%
%


\section{Introduction}
\label{sec:intro}




Remote sensing allows for the collection of information about the earth's surface. For example, satellite-derived nighttime light (NTL), is measured using sensors onboard polar-orbiting satellites, such as the Visible Infrared Imaging Radiometer Suite (VIIRS)~\cite{elvidge2017viirs}. These sensors measure radiance at wavelengths sensitive to artificial light emissions. The raw sensor data undergo calibration and correction procedures to mitigate noise from instrumental and environmental artifacts~\cite{coesfeld2020background}, yielding spatially explicit estimates of light intensity measured in radiance (nW/cm$^2$/sr). 

Our study aims to identify the most effective pre-processing techniques for NTL by systematically evaluating and comparing a range of pre-processing techniques. To our knowledge, the use of two such methods in tandem, built masking with quality filtering and imputation, is novel.  We focus on hurricanes, due to their increasingly strong impact, though the results of this study could be applied more broadly.

Natural disasters, such as hurricanes, can inflict significant economic burdens. The unpredictable recovery period following these disasters can span years, and sometimes regions fail to return to their pre-disaster conditions~\cite{barton}. Between 2010 and 2019, the economic impact of weather and climate events in the U.S. was 131 billion dollars. Only 12 of these events were hurricanes and yet they contributed to over 50\% of the total cost~\cite{noaa2010sdecade}. Previous research has studied the impact of hurricanes on economic activity utilizing NTL data~\cite{ishizawa2017understanding, lin2023critical, xu2021spatial, jia2023estimating, gillespie2007assessment}. However, debates regarding the reliability of NTL data in predicting natural disaster impact persist~\cite{skoufias2021can}. 

We conduct a qualitative case study on Hurricane Michael (2018) and a quantitative correlation analysis including three other hurricanes in which we focus on establishing a robust correlation between loss data and change in NTL pre- and post-hurricane. This is similar to the validation procedure employed in Jia et al. (2023) but directly using damage estimates rather than power outages as a proxy~\cite{minghui}. 
Unique to our study is the zip code level granularity of the data, allowing for a more detailed analysis compared to previous studies that often examined larger geographic areas. Our findings underscore the impact of pre-processing selection on the coherence and reliability of NTL data. We aim to improve methodological consistency and precision in remote sensing analyses of disaster-induced economic losses. This could aid in more accurate damage prediction and better-informed disaster response strategies, ultimately contributing to more resilient communities in the face of natural disasters.

\section{Data}
\label{sec:data}

\subsection{Hurricane Damage Data}
\label{sec:lossdata}
We used modeled damage ratio estimate data from KatRisk, a catastrophe model, for four significant hurricanes within the state of Florida — Irma (2017), Harvey (2017), Michael (2018), and Matthew (2016)~\cite{katrisk}. This model provides percentage damage estimates based on storm reconstructions within each zip code. We categorized zip codes into two groups based on the extent of their losses: highly impacted and minimally impacted. This classification also considers population sizes to ensure comparability, recognizing the relationship between population density and economic losses.

\subsection{Nighttime Lights}
\label{sec:ntl}
\begin{figure}[t!]
  \centering
  \includegraphics[width=0.49\textwidth]{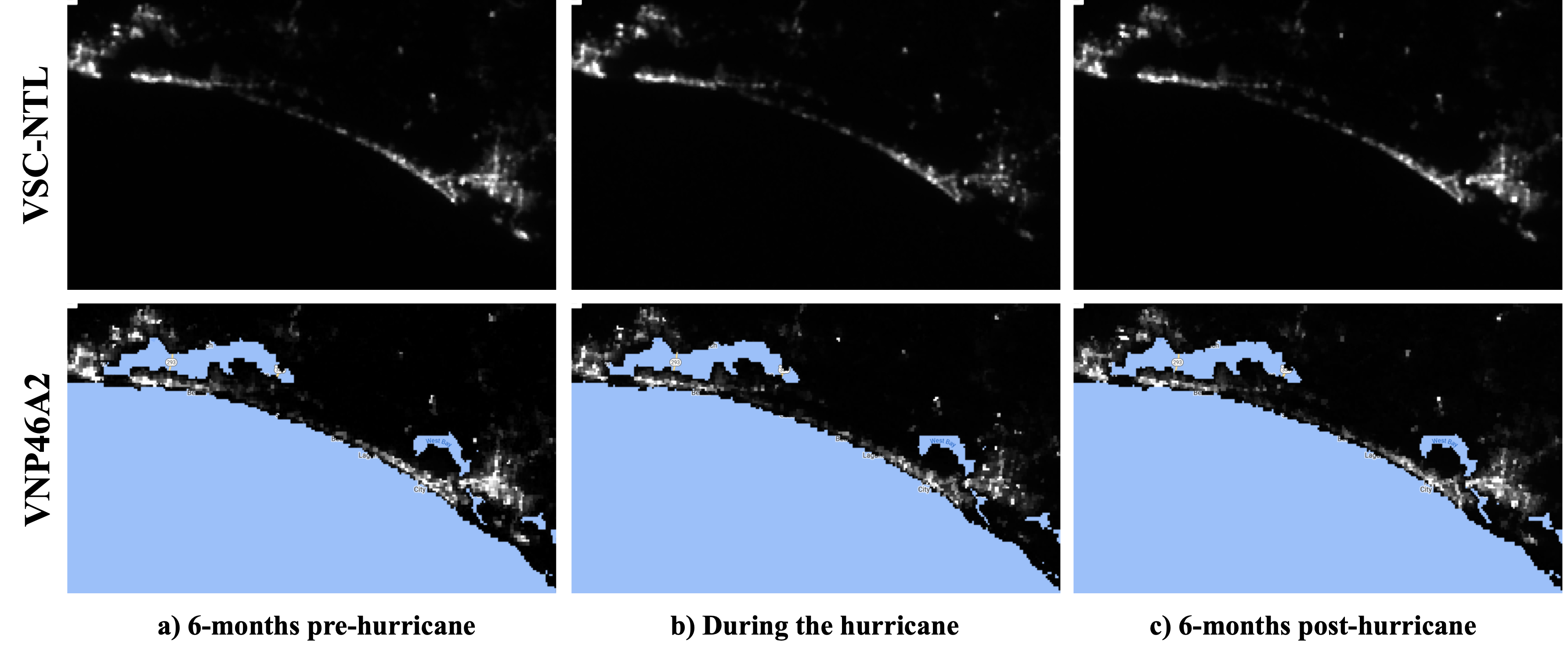}
  \caption{The DNB Radiance variations for two datasets in a Florida Area: Pre-, During, and Post-Hurricane Michael.}
  \label{fig:images}
\end{figure}

Our study employs NTL data from VIIRS, and more specifically, from two datasets: VIIRS Stray Light Corrected Nighttime Day/Night Band (DNB) Composites (VSC-NTL)~\cite{elvidge2017viirs,eog_viirs_ntl,elvidge2013viirs,mills} and VIIRS Lunar Gap-Filled BRDF Nighttime Lights Daily (VNP46A2), which is part of NASA's Black Marble product suite~\cite{roman,gee}. Figure~\ref{fig:images} depicts the DNB radiance for an area in Florida six months pre-hurricane, during the hurricane month, and six months post-hurricane for both datasets in their raw form. There is a noticeable decline in radiance within zip codes 32401 and 32405, corresponding to the bottom right hand corner of the image, during the month of the hurricane. Note that VNP46A2 doesn't include coverage of ocean regions (indicated in blue) while VSC-NTL does.


\subsubsection{VSC-NTL}
\label{sec:viirs}

Prior research using VIIRS DNB suggests that the dataset is noisy~\cite{yuan2019filtering}, especially close to the equator. Therefore, we selected VSC-NTL, which applies a stray light correction algorithm. This dataset has a resolution of 463.83 meters and is available monthly from January 2014 to August 2023. It includes two bands: one for DNB radiance and another for cloud cover information serving as quality control for the DNB band~\cite{mills}. The stray light correction algorithm is intended to mitigate the impact of stray sunlight and moonlight detected by satellite sensors while the cloud cover band attempts to correct errors in DNB radiance due to clouds.


\subsubsection{VNP46A2}
\label{sec:bm}


The VNP46A2 dataset from NASA's Black Marble product provides advanced corrections which further reduce noise in the data. The dataset features lunar irradiance modeling, atmospheric correction, seasonal variation adjustments, and temporal gap-filling to isolate NTL signals from VIIRS DNB data~\cite{roman,martinez2023suitability}. It has a resolution of 500 meters and is available daily from January 2012 to the present (December 2023). 

To enable comparison with VSC-NTL, we aggregate the data to a monthly median. We focus on two of the seven bands: the gap-filled DNB BRDF-corrected NTL band and the quality filter cloud mask band, which respectively measure DNB radiance and specify quality across seven dimensions: day/night, land/water background, cloud mask quality, cloud detection results and confidence indicator, shadow detection, cirrus detection, and snow/ice surface.


\subsection{Dynamic World}
Built masking involves masking NTL imagery by removing all but the built areas, which serve as a proxy for human activity. For this purpose, we leveraged data from the Dynamic World (DW) dataset which uses a pixel-level classification model based on Sentinel-2 to provide land-use/land-cover labels~\cite{dw}. The DW dataset provides near real-time data with a 10-meter spatial resolution. It comprises ten bands, including one that indicates the likelihood that each pixel is associated with a particular land use/cover category, and another that denotes the most probable land use/cover classification for each pixel. 

\section{Methods}

\subsection{Data Extraction}
\label{sec:dataextraction}

A comprehensive dataset was constructed by iterating over impacted zip codes for all hurricanes for each of the two NTL datasets. To capture the environmental changes associated with each event, our analysis required a temporal window extending 12 months before and after the hurricane impact date. For each impacted zip code and each month, we clipped the datasets to the given month and geometric area of the zip code. We then applied a series of pre-processing methods, which we will elaborate upon in Section~\ref{sec:pre-processing}. For each timestep and zip code, we extracted the DNB radiance values from the DNB band and calculated the mean value. By analyzing satellite imagery data for each monthly interval, we were able to paint a detailed picture of the hurricane's immediate and lingering impacts over a 24-month timeline, providing a comprehensive understanding of the environmental and socio-economic changes induced by each natural disaster.

\subsection{Data pre-processing}
\label{sec:pre-processing}

Pre-processing of NTL data is necessary due to the inherent noise associated with instrumentation and environmental factors~\cite{coesfeld2020reducing,man2021normalization}. The choice of pre-processing method is critical, as it underpins the integrity of the signal related to economic activity and, therefore, influences the conclusions drawn from the data. Our pre-processing regimen includes three methods which can be used individually or in tandem: value thresholding; built masking; and quality filtering and imputation. For VSC-NTL, we experimented with every permutation of the three pre-processing techniques, including the option to omit a given method, resulting in twelve unique combinations. Since VNP46A2 already undergoes a noise reduction filter, value thresholding is unnecessary. Hence, for VNP46A2, we tested every combination of the remaining two pre-processing methods, resulting in four unique combinations.

\subsubsection{Value Thresholding}
Value thresholding involves filtering out anomalous DNB radiance values. Following~\cite{barton}, we defined anomalies as values outside the 0-50 nW/cm²/sr range. This approach targets errors caused by background luminance and sensor blooming, which can impair accurate DNB radiance detection. 
Two thresholding approaches were employed: clipping and removal. In the first case, pixel values are clipped to a 0 to 50 range. In the second case, pixel values outside the specified range are removed. 

\subsubsection{Built Masking}
Built masking refers to excluding all pixel values corresponding to non-built areas. We utilized the DW dataset to develop and apply a built mask, retaining only DNB radiance values linked to built structures. This relates to our focus on human-centric activity, under the assumption that illumination inside the built area might contribute more significantly to such activity.

\subsubsection{Quality Filtering and Imputation}
Quality filtering differentiates between low and high-quality pixels, masks the former, and imputes their values. Although the criteria for low-quality pixels vary slightly between VSC-NTL and VNP46A2, the imputation methodology remains consistent. For VSC-NTL, the cloud cover band indicates the count of cloud-free observations per month per pixel; a low count implies reduced reliability of the observed DNB radiance. We designated high-quality pixels for VSC-NTL as those with a non-zero value in the cloud cover band. For VNP46A2, we employed the quality filter cloud bit mask to identify high-quality pixels. We define high-quality pixels as those whose background is composed of land, excluding water and desert, and those that do not contain snow, ice or shadows. Additionally, we require a high cloud mask quality with at least a ``probably clear'' cloud detection confidence, and no cirrus clouds detected.
Pixels not meeting these high-quality criteria for both datasets were masked and their values were imputed using a weighted average of historical high-quality pixel values with weights based on the temporal proximity to the missing observation. 

\section{Results}
\label{sec:results}

\begin{figure}[!htb]
    \centering
    \begin{subfigure}{1.02\columnwidth} 
        \centering
        \includegraphics[scale=0.111]{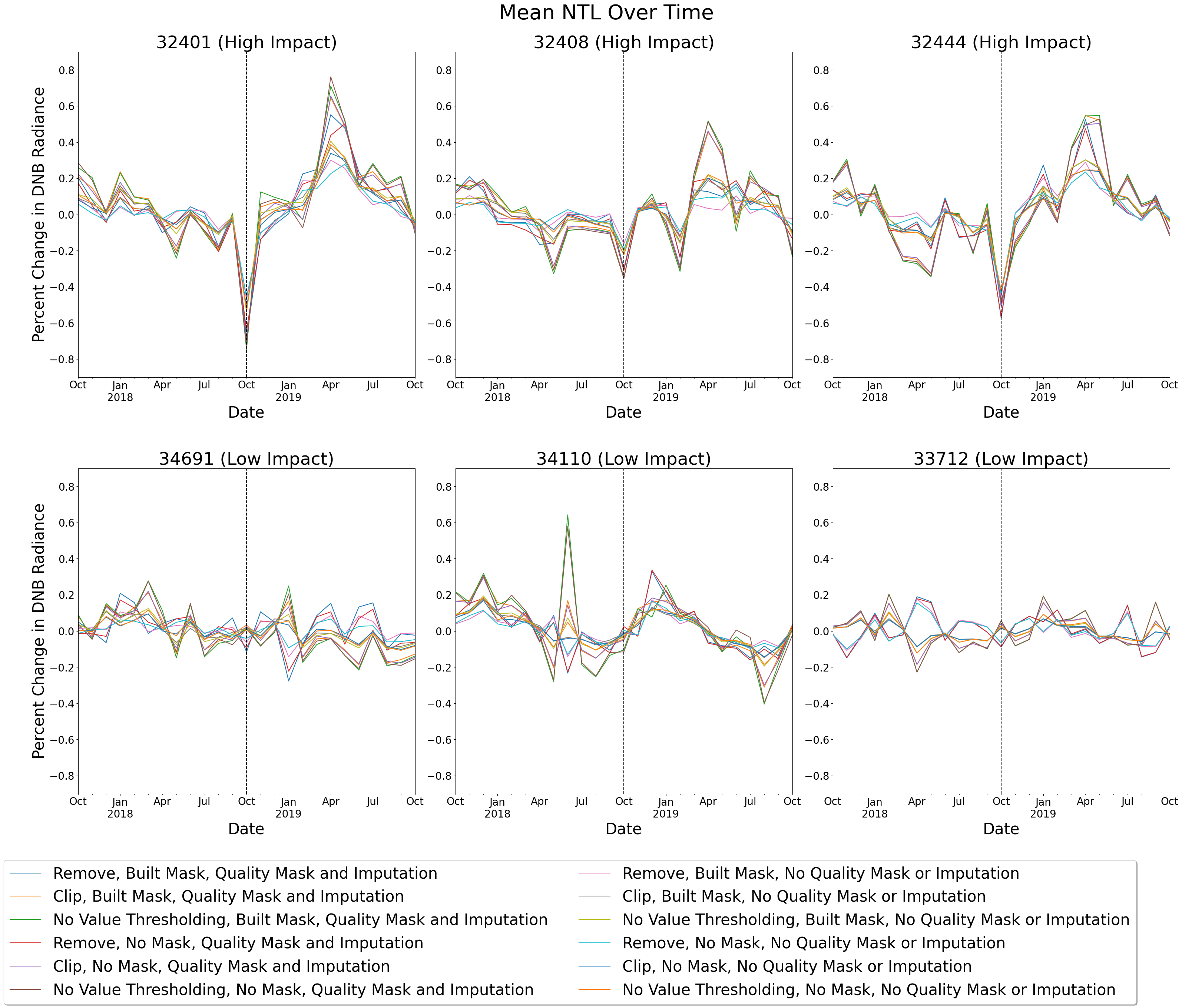}
        \caption{VSC-NTL}
        \label{fig:vsc}
    \end{subfigure}
    
    
    \begin{subfigure}{1.02\columnwidth} 
        \centering
        \includegraphics[scale=0.111]{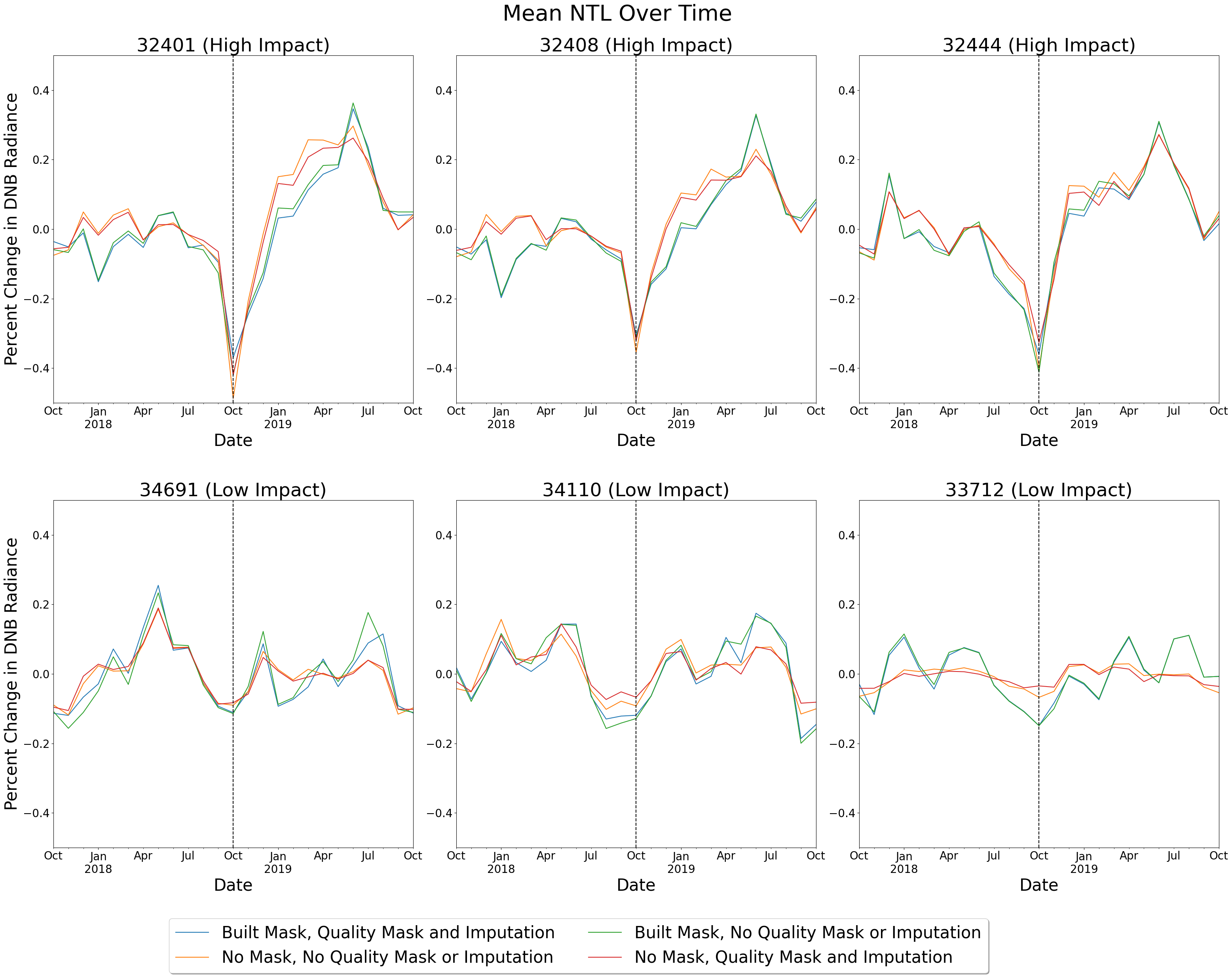}
        \caption{VNP46A2}
        \label{fig:bm}
    \end{subfigure}
    
    \caption{Monthly percentage change in average NTL from October 2017 to 2019 for Hurricane Michael's high/low impacted zip codes. The event date is shown with a vertical line.}
    \label{fig:michael}
\end{figure}

We applied the described pre-processing methods to VSC-NTL and VNP46A2 to investigate the impact of Hurricane Michael, a Category 5 hurricane that struck in October 2018. We focused on zip codes within Florida, categorizing them based on the modelled damage ratios. The three zip codes with the greatest damage ratios and the three with the least were selected for a comparative case study. 
Figure~\ref{fig:michael} shows the percentage change in DNB radiance for both VSC-NTL and VNP46A2 relative to a 6-month rolling average across all relevant pre-processing methods and zip codes. The plot reveals divergent patterns among the pre-processing methods, suggesting that each may lead to different interpretations of Hurricane Michael's immediate and extended impact. Notably, zip codes with higher damage exhibited a more pronounced decline in NTL during the hurricane, highlighting the potential relationship between radiance change and economic damage. Such trends underscore the utility of NTL data in assessing and quantifying the economic consequences of major natural disasters. 


\label{sec:corrlosses}

To establish the most effective dataset and pre-processing technique for utilizing NTL radiance as a proxy for economic activity, we calculated the Pearson correlation coefficient (PCC) between the decrease in DNB radiance during hurricane events and the damage ratios in the hurricane damage data. Higher correlation coefficients are indicative of a stronger association, suggesting that certain pre-processing methods perform better in denoising the data than others. Our study included data from four notable hurricanes that impacted the state of Florida — Michael, Matthew, Irma, and Harvey - and focused on 25 zip codes that experienced substantial damage, excluding those with a damage ratio below one percent. The outcomes of this process are summarized in Table 1, which enumerates the correlation coefficients for each pre-processing method within VSC-NTL and VNP46A2. The pre-processing methods are denoted as follows: clipping, which adjusts outlier values to within a specified range; removal, which excludes anomalous data points; built masking, which isolates radiance measurements from built pixels; and quality filtering and imputation, which refines the data quality and supplements missing values with computed historical data. The results show that within VSC-NTL, the clipping approach alone corresponds to the highest correlation with economic damage. Conversely, for VNP46A2, quality filtering and imputation stand out as the most closely related to economic damage.


\begin{table}[htbp]
    \centering
    \begin{tabular}{p{1.6cm}p{4.8cm}c}
        \textbf{Dataset} & \textbf{Pre-processing Method} & \textbf{PCC} \\
        \hline
        \multirow{12}{*}{VSC-NTL} & Raw & 0.549 \\
         & Clip & 0.553 \\
         & Remove & 0.472 \\
         & Built mask & 0.435 \\
         & Clip, Built mask & 0.446 \\
         & Remove, Built mask & 0.333 \\
         & Quality filter & 0.400 \\
         & Clip, Quality filter & 0.432 \\
         & Remove, Quality filter & 0.337 \\
         & Built mask, Quality filter & 0.331 \\
         & Clip, Built mask, Quality filter & 0.361 \\
         & Remove, Built mask, Quality filter & 0.295 \\
        \hline
        \multirow{4}{*}{VNP46A2} & Raw & 0.738 \\
         & Built mask & 0.559 \\
         & Quality filter & \textbf{0.761} \\
         & Built mask, Quality filter & 0.703 \\
        \hline
    \end{tabular}
    \caption{Correlation between the percentage drop in NTL and damage ratio for hurricanes Michael, Matthew, Irma, and Harvey.}
    \label{tab:drops}
\end{table}

\section{Discussion}
Our study presents a novel analysis of NTL radiance data to assess the economic impact of hurricanes, using two distinct datasets and various pre-processing techniques. Figure~\ref{fig:michael} underscores the importance of methodological choices, such as using different datasets or pre-processing methods, in NTL data analysis for disaster response. For example, VNP46A2 displays less variability and smoother trends while VSC-NTL, particularly in its raw form, shows a lot of variability, which might obscure the signal. Oscillations in NTL radiance outside the disaster window are expected and could be due to other natural disasters or changes in human behavior and activity. Figure~\ref{fig:michael} also highlights the potential for NTL intensity changes to act as an indicator of economic damage and recovery. For example, the impact of Hurricane Michael on NTL in high-impact versus low-impact zip codes is visibly different. Finally, we get a sense for the pattern of recovery from disasters and the differential resilience~\cite{huang2019human} and effectiveness of recovery efforts across communities. In future work, we may consider the effect of government aid on observed data.

As shown in Table~\ref{tab:drops}, all of the pre-processing methods involving VNP46A2 outperformed those involving VSC-NTL. This is unsurprising as VNP46A2 undergoes additional corrective procedures not performed on VSC-NTL. Our hypothesis that examining only built areas, as the hubs of human activity, would eliminate much of the noise and show significant changes in light emissions close to the true signal of the change in economic activity was not corroborated. This may suggest that filtering for only built terrain does not serve as a good proxy for human activity filtering or it may point to minor discrepancies in the land-use/land-cover labels. We hope to examine this phenomenon more in future studies.
A robust correlation (0.761), however, was found with VNP46A2 when processed with quality masking and imputation. This suggests that leveraging the quality flags and our imputation strategy is a good way to extract true signal from NTL data. This makes sense since VNP46A2 offers a range of quality flags which can be used to refine the data.

In future work, we would like to forecast post-hurricane NTL time series, providing insights into the trajectory of recovery or further decline in affected areas. Our goal is to refine the precision of disaster impact assessments and to enhance the predictive capabilities of NTL analysis. By doing so, we hope to contribute to more effective disaster preparedness and recovery strategies, mitigating the long-term economic and social consequences of such catastrophic events.

\bibliographystyle{IEEEbib}
\bibliography{strings,refs}

\section{Acknowledgments}
The authors would like to thank Dr. Zhen Zeng and Dr. Elizabeth Fons for providing technical consulting. 

This paper was prepared for informational purposes “in part” by the Artificial Intelligence Research group of JPMorgan Chase \& Co and its affiliates (“J.P. Morgan”) and is not a product of the Research Department of J.P. Morgan.  J.P. Morgan makes no representation and warranty whatsoever and disclaims all liability, for the completeness, accuracy or reliability of the information contained herein.  This document is not intended as investment research or investment advice, or a recommendation, offer or solicitation for the purchase or sale of any security, financial instrument, financial product or service, or to be used in any way for evaluating the merits of participating in any transaction, and shall not constitute a solicitation under any jurisdiction or to any person, if such solicitation under such jurisdiction or to such person would be unlawful.  


\end{document}